\documentclass{article}
\usepackage{spconf,amsmath,graphicx}
\usepackage{amsfonts}
\usepackage{amsmath}
\usepackage{adjustbox}
\usepackage{algorithmicx}
\usepackage{algpseudocode}
\usepackage[ruled]{algorithm}
\usepackage{graphicx}
\usepackage{caption}
\usepackage{graphicx,times,amsmath} 
\usepackage{caption}
\usepackage{subcaption}
\usepackage[rightcaption]{sidecap}
\usepackage{caption}
\usepackage{multirow}
\usepackage{hhline}
\usepackage{amsmath}
\usepackage{epstopdf}
\usepackage{tabularx}
\usepackage{tikz}
\usetikzlibrary{matrix}
\usepackage{mathtools, nccmath}
\usepackage{pst-node}

\graphicspath{ {images/} }

\usepackage{alphalph}

\title{Ising-Dropout: A Regularization Method for Training and\\ Compression of Deep Neural Networks}

\name{Hojjat~Salehinejad and Shahrokh~Valaee}
\address{Department of Electrical \& Computer Engineering, University of Toronto, Toronto, Canada \\
\textit{hojjat.salehinejad@mail.utoronto.ca, valaee@ece.utoronto.ca}}

\usepackage{fancyhdr}
\fancypagestyle{pageStyleOne}{%
\normalfont\footnotesize
    \fancyhf{}
\fancyhead[L]{\fontsize{8}{8} \selectfont This paper is accepted at $44^{th}$ IEEE International Conference on Acoustics, Speech and Signal Processing (IEEE ICASSP), 2019.}
}

\begin{document}
\newcommand*{\img}{%
  \includegraphics[
    width=\linewidth,
    height=20pt,
    keepaspectratio=false,
  ]{example-image-a}%
}

\maketitle
\thispagestyle{pageStyleOne}

\begin{abstract}

Overfitting is a major problem in training machine learning models, specifically deep neural networks. This problem may be caused by imbalanced datasets and initialization of the model parameters, which conforms the model too closely to the training data and negatively affects the generalization performance of the model for unseen data. The original dropout is a regularization technique to drop hidden units randomly during training. In this paper, we propose an adaptive technique to wisely drop the visible and hidden units in a deep neural network using Ising energy of the network. The preliminary results show that the proposed approach can keep the classification performance competitive to the original network while eliminating optimization of unnecessary network parameters in each training cycle. The dropout state of units can also be applied to the trained (inference) model. This technique could compress the number of parameters up to $41.18\%$ and $55.86\%$ for the classification task on the MNIST and Fashion-MNIST datasets, respectively.
\end{abstract}
\begin{keywords}
Compressed neural networks, dropout, Ising model, overfitting, training deep neural networks.
\end{keywords}
\section{Introduction}
\label{sec:intro}

Neural networks are constructed from layers of activation function, which produce a value by optimizing a set of weights~\cite{salehinejad2017recent}. This complicated connection between the weights of a network, if trained well and enough data is available, can model complex systems. The wider and deeper a network is, the more computational time is needed to optimize the weights. However, in real world problems, most of datasets are imbalanced and limited quantities are available; for example fraud transaction versus healthy transaction in a bank or rare diseases in medical imaging~\cite{salehinejad2018image},\cite{salehinejad2018generalization}. This problem may result in overfitting in training neural networks and the model may not be generalized. A variety of regularization methods have been developed to reduce overfitting, including early-stopping~\cite{salehinejad2017recent}, adding weight penalties in the cost function of the networks such as $L_{1}$ and $L_{2}$~\cite{zaremba2014recurrent}, and dropout~\cite{srivastava2014dropout}. 

Dropout is a very effective regularization technique for training neural networks~\cite{srivastava2014dropout}. This approach drops a random set of units and corresponding connection from the network during training and uses all the units at the inference (test) time. This method not only reduces the number of parameters to optimize in each training iteration, but also prevents units from too much co-adaptation~\cite{srivastava2014dropout}. A neural network with $n$ units can be seen as a set of $2n$ small (thinned~\cite{srivastava2014dropout}) networks. Therefore, the maximum number of parameters is $O(n^{2})$. Dropout selects a network from this set of parameters at each training iteration for optimization. Since the weights of thinned networks are shared, a subset of parameters is updated at each training iteration. However, since the number of possible thinned networks is of exponential order, it is not feasible to update all networks~\cite{srivastava2014dropout}. 

Ising model is widely used for modeling phenomena in physics such as working of magnetic material~\cite{kadowaki1998quantum}. In this paper, we propose using Ising energy~\cite{kadowaki1998quantum} to model dropout in deep neural networks. We map activation values of each single neuron to a cost value (Ising weight) in the Ising model. The Ising weights are shipped to an optimizer, an accelerated hardware architecture designed for solving combinatorial optimization problems using Markov-chain Monte-
Carlo (MCMC) search~\cite{matsubara2017ising}, to minimize the cost (energy) of connections by flipping the binary state variables of the units. The generated state variable is then applied as a mask on the weight tensors for backpropagation and inference. This process is conducted for every mini-batch of training data.

\section{Proposed Method}
We propose an adaptive solution compared to random dropout using Ising model~\cite{matsubara2017ising} for training deep multilayer perceptron (MLP) networks. 

\begin{figure}[!t]
\centering
\captionsetup{font=small}
\includegraphics[width=0.4\textwidth]{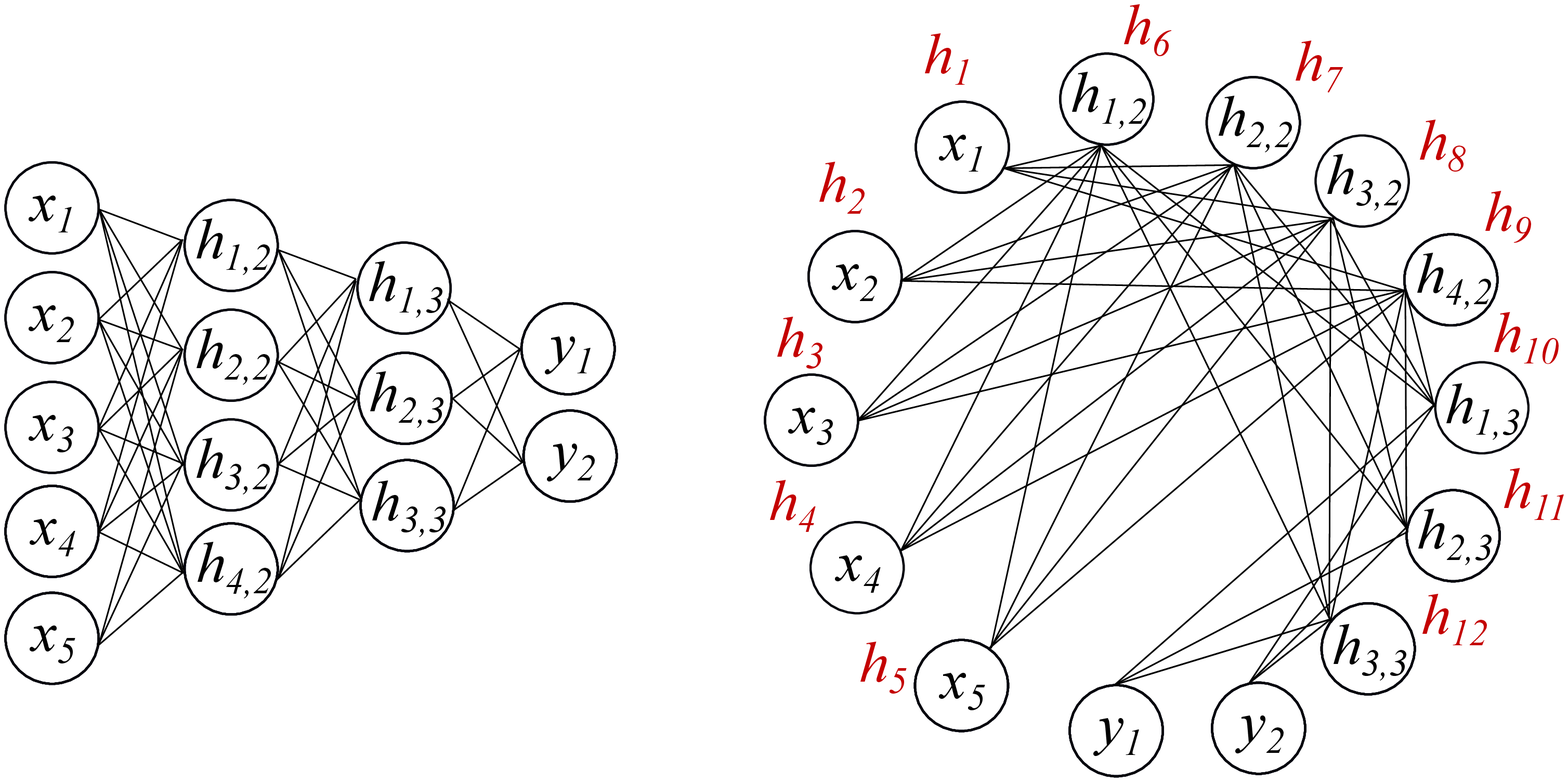}   
\caption{Left: An MLP network with 5 inputs, two hidden layers, and two outputs; Right: Representing the MLP in left as subset of a fully connected graph. The candidate nodes for dropout are labeled in red.}
\label{fig:ising_neuralgraph}
\vspace{-2mm}
\end{figure}
\alglanguage{pseudocode}
\begin{algorithm}[!tp]
\footnotesize
\caption{Ising-Dropout}
\begin{algorithmic}
\State Initialize Weights $\mathbf{W}$
\State Initialize Masks $\mathbf{M}=1$ 
\State Initialize Loss as $L$
\For {$t=0 \to T$} // iteration counter
\For {$i=0 \to I$} // mini-batch counter
\State Load mini-batch $(\mathbf{X},\mathbf{y})$
\If {$t==0 \:\& \: i==0$} 
\State $\mathbf{W}, L =$ backPropagation($\mathbf{X},\mathbf{y},\mathbf{M}$)
\State $\mathbf{W^{*}}=\mathbf{W}$ // a copy of $\mathbf{W}$
\Else
\State $\mathbf{W}=$ backPropagation($\mathbf{X},\mathbf{y},\mathbf{\bar{M}}$)
\State $\mathbf{W^{*}}=\mathbf{W}\times \mathbf{\bar{M}}+\mathbf{W^{*}}\times$NOT$(\mathbf{\bar{M}})$
\State $L=$inference$(\mathbf{X},\mathbf{y},\mathbf{M^{*}})$ // compute loss
\EndIf
\State $\mathbf{s}=$Ising-Dropout($\mathbf{W^{*}}$) // perform dropout
\State $\mathbf{\bar{M}}=1$
\For {$j=1 \to N$} // each candidate node to drop
\If {$\mathbf{s}[j]==0$}
\State $\mathbf{\bar{M}}[j]=0$
\EndIf
\EndFor 

\EndFor 
\EndFor
\end{algorithmic}
\label{alg:da}
\end{algorithm}
\alglanguage{pseudocode}

\subsection{Model Architecture}
\begin{figure}[!t]
\centering
\captionsetup{font=small}
\includegraphics[width=0.3\textwidth]{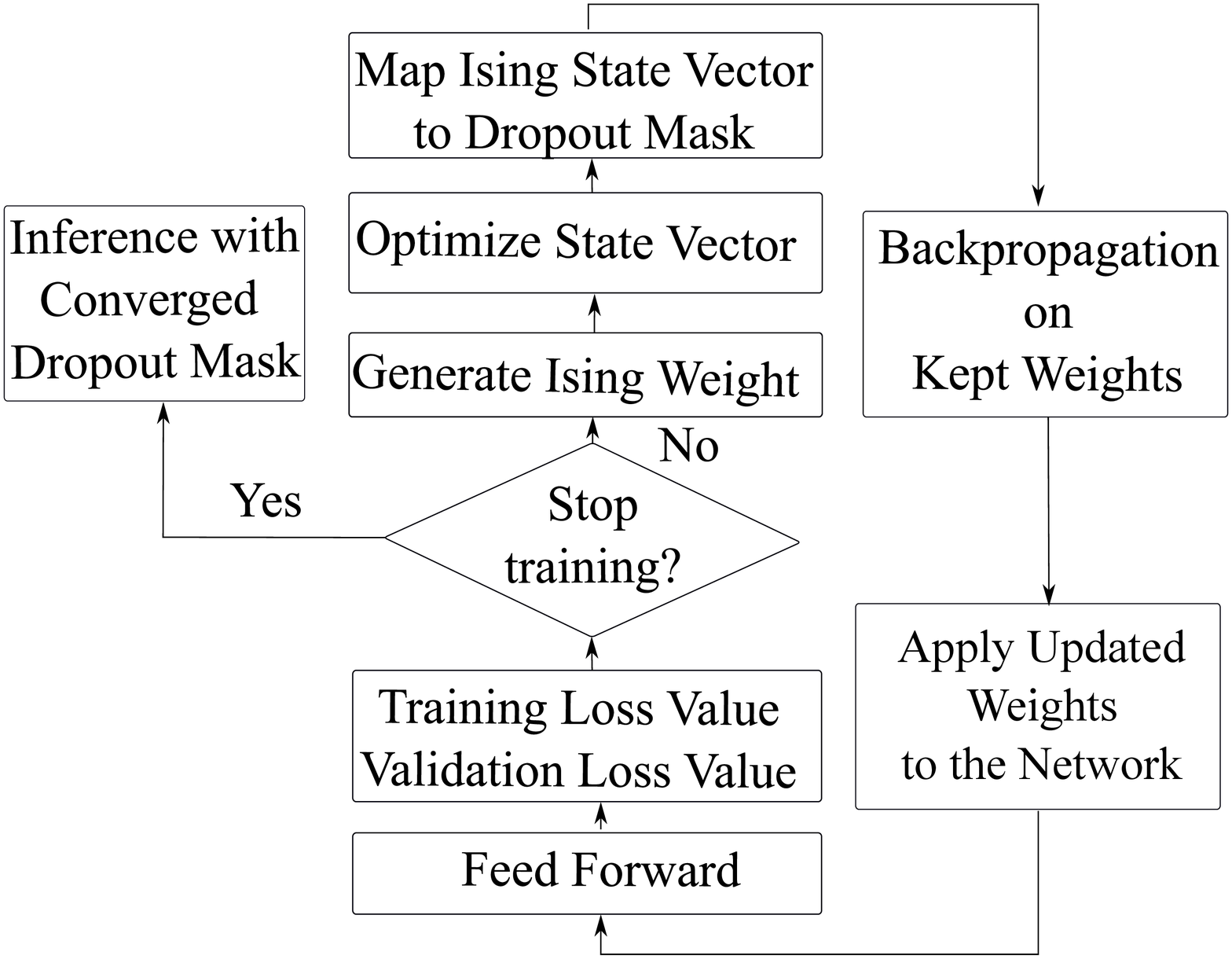}   
\caption{Dropout using Ising model.}
\label{fig:ising_algorithm}
\end{figure}

We consider an MLP network as a subgraph of a fully connected graph, where each candidate node for dropout is indexed as $h_{i}$ as in Figure~\ref{fig:ising_neuralgraph}. Figure~\ref{fig:ising_algorithm} shows the overall system design of training a neural network with Ising-Dropout. 
Since the Ising model optimization is a combinatorial NP-hard~\cite{matsubara2017ising} problem, we use the Fujitsu Digital Annealer (DA)~\cite{matsubara2017ising}. The DA machine performs an optimization process for each training epoch of the neural network and generates a state variable for the network weights. 

The pseudocode of training procedure is illustrated in Algorithm~\ref{alg:da}. For the first iteration over a mini-batch in training, the backpropagation is performed on the randomly initialized weights $\mathbf{W}$ of the network. The updated weights after backpropagation are then mapped to a cost matrix for Ising-Dropout as described in the next subsection. The returned state vector $\mathbf{s}$ is translated to a set of matrices $\mathbf{\bar{M}}$ to be applied as a mask on the weights of the network. This process will repeat for a number of iterations or will be stopped using early-stopping~\cite{salehinejad2017recent}.

\begin{figure}[!tp]
\centering
\captionsetup{font=small}
          \begin{subfigure}[t]{0.24\textwidth}
        \centering
        \captionsetup{font=small}
                \includegraphics[width=0.98\textwidth]{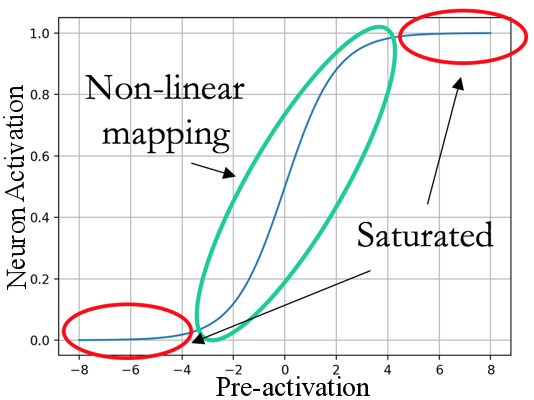}
                \caption{Neuron activation using Sigmoid function.}
                \label{fig:radial_sampling}
        \end{subfigure}%
                        ~        
                      \begin{subfigure}[t]{0.24\textwidth}
                   \captionsetup{font=small}
        \centering
                \includegraphics[width=0.98\textwidth]{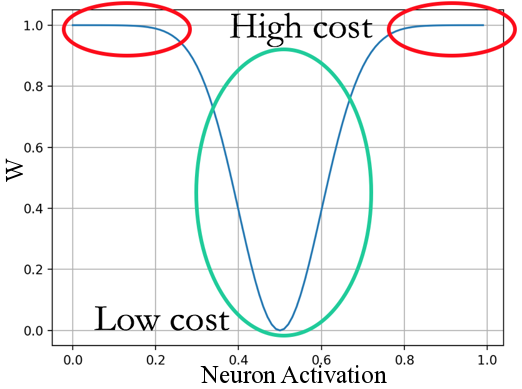}
                \caption{Mapping neuron activation to Ising model weight.}
                \label{fig:radial_sampling_angular}
        \end{subfigure}%
\caption{Distribution of Sigmoid activation values for different pre-activations. The activation value is then mapped to a cost value (weight) for the Ising model.}
\label{fig:sigmoid_ising_mapping}
\end{figure}


\begin{table*}[!h]
\centering
\captionsetup{font=footnotesize}
\caption{ Performance comparison of various dropout method on MNIST dataset. $h_{i}$: the percentage of dropped units for layer $h_{i}$; P: total number of parameters in the network. Acc: test set classification accuracy. The size of each layer in order of stacking is in parenthesis under network layers. Training refers to applying dropout only in training phase and training+inference refers to applying dropout to training and test (inference) phases.}
\begin{adjustbox}{width=1\textwidth}
\begin{tabular}{|c|c|c|c|c|c||c|c|c|c|c|c||c|c|c|c|c|c|c|}
\hline
Network Layers                                                                                      & \multicolumn{5}{c||}{(784,100,100,10)}                                                                       & \multicolumn{6}{c||}{(784,100,50,50,10)}                                                                     & \multicolumn{7}{c|}{(784,100,50,50,25,10)}                                                                                                              \\ \hline
\multirow{2}{*}{Model}                                                                              & \multicolumn{4}{c|}{\begin{tabular}[c]{@{}c@{}}Dropout Rate\\ P=89,610\end{tabular}} & \multirow{2}{*}{Acc} & \multicolumn{5}{c|}{\begin{tabular}[c]{@{}c@{}}Dropout Rate\\ P=86,610\end{tabular}} & \multirow{2}{*}{Acc} & \multicolumn{6}{c|}{\begin{tabular}[c]{@{}c@{}}Dropout Rate\\ P=87,635\end{tabular}}                                     & \multirow{2}{*}{Acc}         \\ \cline{2-5} \cline{7-11} \cline{13-18}
                                                                                                    & \multicolumn{1}{c|}{$h_{0}$}        & $h_{1}$        & $h_{2}$       & Total         &                      & \multicolumn{1}{c|}{$h_{0}$}     & $h_{1}$    & $h_{2}$    & $h_{3}$    & Total      &                      & \multicolumn{1}{c|}{$h_{0}$} & $h_{1}$ & $h_{2}$ & $h_{3}$ & \multicolumn{1}{c|}{$h_{4}$} & Total                        &                              \\ \hline
No Dropout                                                                                          & 0\%                                 & 0\%            & 0\%           & 0\%           & 94.65\%              & 0\%                              & 0\%        & 0\%        & 0\%        & 0\%        & 95.02\%              & 0\%                          & 0\%     & 0\%     & 0\%     & \multicolumn{1}{c|}{0\%}     & 0\%                          & 94.40\%                      \\ \hline
Dropout (p=0.5)                                                                                     & 0\%                                 & 50.00\%        & 50.00\%       & 06.26\%       & 91.02\%              & 0\%                              & 50.00\%    & 50.00\%    & 50.00\%    & 04.74\%    & 87.59\%              & 0\%                          & 50.00\% & 50.00\% & 50.00\% & \multicolumn{1}{c|}{50.00\%} & 05.27\%                      & 56.89\%                      \\ \hline
\begin{tabular}[c]{@{}c@{}}Dropout (p=0.5)  \\ (input layer included)\end{tabular}                  & 50.00\%                             & 50.00\%        & 50.00\%       & 50.00\%       & 85.08\%              & 50.00\%                          & 50.00\%    & 50.00\%    & 50.00\%    & 50.00\%    & 82.05\%              & 50.00\%                      & 50.00\% & 50.00\% & 50.00\% & 50.00\%                      & \multicolumn{1}{l|}{50.00\%} & \multicolumn{1}{l|}{64.03\%} \\ \hline
Ising-Dropout (training)                                                                            & 0\%                                 & 38.62\%        & 42.43\%       & 04.88\%       & 93.83\%              & 0\%                              & 49.21\%    & 47.37\%    & 26.37\%    & 04.47\%    & 93.78\%              & 0\%                          & 42.59\% & 46.62\% & 43.18\% & \multicolumn{1}{c|}{51.25\%} & 04.64\%                      & 90.15\%                      \\ \hline
\begin{tabular}[c]{@{}c@{}}Ising-Dropout (training)\\ (input layer included)\end{tabular}           & 38.60\%                             & 32.18\%        & 25.15\%       & 37.71\%       & 93.47\%              & 40.21\%                          & 33.00\%    & 38.31\%    & 26.43\%    & 39.64\%    & 90.72\%              & 42.18\%                      & 31.78\% & 33.18\% & 37.00\% & 25.37\%                      & \multicolumn{1}{l|}{41.18\%} & \multicolumn{1}{l|}{90.28\%} \\ \hline
\begin{tabular}[c]{@{}c@{}}Ising-Dropout \\(training+inference)                                                                 \end{tabular} & 0\%                                 & 38.62\%        & 42.43\%       & 04.88\%       & 92.10\%              & 0\%                              & 49.21\%    & 47.37\%    & 26.37\%    & 04.47\%    & 91.42\%              & 0\%                          & 42.59\% & 46.62\% & 43.18\% & 51.25\%                      & 04.64\%                      & \multicolumn{1}{l|}{91.54\%} \\ \hline
\begin{tabular}[c]{@{}c@{}}Ising-Dropout \\(training+inference)\\ (input layer included)\end{tabular} & 38.60\%                             & 32.18\%        & 25.15\%       & 37.71\%       & 91.40\%              & 40.21\%                          & 33.00\%    & 38.31\%    & 26.43\%    & 39.64\%    & 90.85\%              & 42.18\%                      & 31.78\% & 33.18\% & 37.00\% & 25.37\%                      & \multicolumn{1}{l|}{41.18\%} & \multicolumn{1}{l|}{90.74\%} \\ \hline
\end{tabular}
\end{adjustbox}
\label{T:mnist_da}
\end{table*}

\begin{table*}[!h]
\centering
\captionsetup{font=footnotesize}
\caption{ Performance comparison between various dropout method on the Fashion-MNIST dataset. $h_{i}$: the percentage of dropped units for layer $h_{i}$; P: total number of parameters in the network. Acc: test set classification accuracy. The size of each layer in order of stacking is in parenthesis under network layers. Training refers to applying dropout only in training phase and training+inference refers to applying dropout to training and test (inference) phases.}
\begin{adjustbox}{width=1\textwidth}
\begin{tabular}{|c|c|c|c|c|c||c|c|c|c|c|c||c|c|c|c|c|c|c|}
\hline
Network Layers                                                                                      & \multicolumn{5}{c||}{(784,100,100,10)}                                                                       & \multicolumn{6}{c||}{(784,100,50,50,10)}                                                                     & \multicolumn{7}{c|}{(784,100,50,50,25,10)}                                                                                                              \\ \hline
\multirow{2}{*}{Model}                                                                              & \multicolumn{4}{c|}{\begin{tabular}[c]{@{}c@{}}Dropout Rate\\ P=89,610\end{tabular}} & \multirow{2}{*}{Acc} & \multicolumn{5}{c|}{\begin{tabular}[c]{@{}c@{}}Dropout Rate\\ P=86,610\end{tabular}} & \multirow{2}{*}{Acc} & \multicolumn{6}{c|}{\begin{tabular}[c]{@{}c@{}}Dropout Rate\\ P=87,635\end{tabular}}                                     & \multirow{2}{*}{Acc}         \\ \cline{2-5} \cline{7-11} \cline{13-18}
                                                                                                    & \multicolumn{1}{c|}{$h_{0}$}        & $h_{1}$        & $h_{2}$       & Total         &                      & \multicolumn{1}{c|}{$h_{0}$}     & $h_{1}$    & $h_{2}$    & $h_{3}$    & Total      &                      & \multicolumn{1}{c|}{$h_{0}$} & $h_{1}$ & $h_{2}$ & $h_{3}$ & \multicolumn{1}{c|}{$h_{4}$} & Total                        &                              \\ \hline
No Dropout                                                                                          & 0\%                                 & 0\%            & 0\%           & 0\%           & 84.24\%              & 0\%                              & 0\%        & 0\%        & 0\%        & 0\%        & 83.48\%              & 0\%                          & 0\%     & 0\%     & 0\%     & \multicolumn{1}{c|}{0\%}     & 0\%                          & 81.87\%                      \\ \hline
Dropout (p=0.5)                                                                                     & 0\%                                 & 50.00\%        & 50.00\%       & 06.26\%       & 77.27\%              & 0\%                              & 50.00\%    & 50.00\%    & 50.00\%    & 04.74\%    & 68.74\%              & 0\%                          & 50.00\% & 50.00\% & 50.00\% & \multicolumn{1}{c|}{50.00\%} & 05.27\%                      & 48.65\%                      \\ \hline
\begin{tabular}[c]{@{}c@{}}Dropout (p=0.5)  \\ (input layer included)\end{tabular}                  & 50.00\%                             & 50.00\%        & 50.00\%       & 50.00\%       & 74.33\%              & 50.00\%                          & 50.00\%    & 50.00\%    & 50.00\%    & 50.00\%    & 64.25\%              & 50.00\%                      & 50.00\% & 50.00\% & 50.00\% & 50.00\%                      & \multicolumn{1}{l|}{50.00\%} & \multicolumn{1}{l|}{49.29\%} \\ \hline
Ising-Dropout (training)                                                                            & 0\%                                 & 49.34\%        & 48.44\%       & 03.62\%       & 82.73\%              & 0\%                              & 44.84\%    & 55.18\%    & 40.75\%    & 04.53\%    & 80.57\%              & 0\%                          & 41.35\% & 39.20\% & 46.06\% & \multicolumn{1}{c|}{45.46\%} & 03.10\%                      & 67.32\%                      \\ \hline
\begin{tabular}[c]{@{}c@{}}Ising-Dropout (training)\\ (input layer included)\end{tabular}           & 44.84\%                             & 49.02\%        & 42.53\%       & 45.07\%       & 85.23\%              & 38.22\%                          & 37.87\%    & 34.00\%    & 32.43\%    & 38.04\%    & 68.42\%              & 56.65\%                      & 45.25\% & 48.31\% & 44.62\% & 48.12\%                      & \multicolumn{1}{l|}{55.86\%} & \multicolumn{1}{l|}{66.36\%} \\ \hline
\begin{tabular}[c]{@{}c@{}}Ising-Dropout \\(training+inference)                                                                 \end{tabular} & 0\%                                 & 49.34\%        & 48.44\%       & 03.62\%       & 83.73\%              & 0\%                              & 44.84\%    & 55.18\%    & 40.75\%    & 04.53\%    & 82.65\%              & 0\%                          & 41.35\% & 39.20\% & 46.06\% & 45.46\%                      &                 03.10\%             & \multicolumn{1}{l|}{73.22\%} \\ \hline
\begin{tabular}[c]{@{}c@{}}Ising-Dropout \\(training+inference)\\ (input layer included)\end{tabular} & 44.84\%                             & 49.02\%        & 42.53\%       & 45.07\%       & 86.21\%              & 38.22\%                          & 37.87\%    & 34.00\%    & 32.43\%    & 38.04\%    & 79.82\%              & 56.65\%                      & 45.25\% & 48.31\% & 44.62\% & 48.12\%                      & \multicolumn{1}{l|}{55.86\%} & \multicolumn{1}{l|}{76.03\%} \\ \hline
\end{tabular}
\end{adjustbox}
\label{T:fashion_da}
\end{table*}

\subsection{Ising Model for Dropout}
 If a neuron's activation value is in the saturated areas, as in Figure~\ref{fig:sigmoid_ising_mapping}(a), it may increase the risk of overfitting. Therefore, the objective is to keep the activation value of a neuron in the non-linear area. That might be a reason why rectified linear units (ReLU)~\cite{nair2010rectified} generally work better than the Sigmoid function, since no upper boundary is defined in the activation function. The weight between node $i$ and $j$ is defined as $w_{i,j}$, the input is a vector $\mathbf{x} = (x_{1}, x_{2}, ..., x_{T})$ and the output is a vector ${\mathbf{y} = (y_{1}, y_{2}, ..., y_{K})}$ where $T$ is the number of inputs and $K$ is the number of data classes.
The Ising cost value for each connection $i,j$ from layer $l-1$ to $l$ is defined as
\begin{equation}
\bar{\gamma}^{(l)}_{i,j}=G(\hat{h}^{(l)}_{i,j}|\mu,\sigma^{'2}),
\end{equation}
such that
\begin{equation}
G(\hat{h}^{(l)}_{i,j}|\mu,\sigma^{'2})=1-e^{-\frac{(\hat{h}^{(l)}_{i,j}-\mu)^2}{2\sigma^{'2}}},
\end{equation}
where $\mu=0.5, \sigma^{'2}=0.01$. The activation value $\hat{h}^{(l)}_{i,j}$ is defined as
\begin{equation}
\hat{h}^{(l)}_{i,j}=\sigma(\frac{1}{Q}\sum_{q=1}^{Q} \bar{h}_{i_{(q)}}^{(l-1)}w^{(l)}_{i,j}).
\end{equation}
where $Q$ is the mini-batch size and $\sigma(\cdot)$ is the Sigmoid activation function. The $\bar{h}_{i}^{(l)}$ activation value is defined as 
\begin{equation}
\bar{h}_{i}^{(l)} = \sigma(\sum_{u=1}^{|h^{(l-1)}|} h^{(l-1)}_{u}w^{(l-1)}_{u,i} + b_{i}^{(l)}) \:\forall \: l\in\{1,...,N-1\},
 \end{equation}
where $\bar{h}_{i}^{(0)}=x_{i}$ and $|h^{(l-1)}|$ is the number (cardinality) of units in layer $l-1$. This cost function is a non-linear mapper from input signal to an output cost value as in Figure~\ref{fig:sigmoid_ising_mapping}(b). This function penalizes saturated neuron activation values by allocating a large cost value. Note that $\bar{\gamma}_{i,j}=0$ if no connection exists between units $i$ and $j$.

The Ising model has a binary state vector where each value represents the state of a unit (0 means dropped) such as $\mathbf{s}=(s_{1},s_{2},...,s_{U})$ which is initialized to 1. The Ising energy model is defined as
\begin{equation}
E(\mathbf{s})=-\sum_{u,v=1}^{U}\gamma_{u,v}s_{u}s_{v}-\sum_{u=1}^{U}b_{u}s_{u} 
\label{eq:ising_energy}
\end{equation}
where $\gamma_{u,v}=sgn(w_{i,j}^{(l)})\bar{\gamma}_{i,j}^{(l)}$ for a given $l$ as in Figure~\ref{fig:ising_neuralgraph} such that $i=u-\sum_{l'=0}^{l-1}|h^{(l')}|$, $j=v-\sum_{l'=0}^{l}|h^{(l')}|$, $b_{u}$ is the bias value of the unit $u$, and $sgn(\cdot)$ is the sign function. 
The binary state vector $\mathbf{s}$ represents dropout state of candidate units. More details about DA and optimization procedure is in~\cite{matsubara2017ising}.

\section{Experiments}
\label{sec:experiments}
Many adaptive dropout methods have been proposed in the literature~\cite{ba2013adaptive},~\cite{wager2013dropout}. The objective in this paper is to study the performance of Ising-Dropout as a regularization method for training deep neural networks and compression of inference model and its affect on the inference performance. The current version of the Fujitsu DA machine has 1,024 state variables. Therefore, we had to limit the size of our models to accommodate the DA. We performed the experiments using MLP networks with various number of hidden layers.

\subsection{Data}
We investigated performance of the proposed method by addressing the classification problem on MNIST~\cite{lecun2010mnist} and Fashion-MNIST~\cite{fashion_mnist} datasets. The MNIST dataset has 10 classes of hand written digits. The Fashion dataset has 10 classes of various clothing items.
The training set had 60,000 samples, which we deployed only 32 epochs over mini-batches to accelerate the training. The samples were shuffled in each training iteration. The test set had 10,000 examples.
\begin{figure*}[!htp]
\centering
\captionsetup{font=small}
          \begin{subfigure}[t]{0.083\textwidth}
        \centering
        \captionsetup{font=small}
                \includegraphics[width=0.8\textwidth]{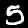}
                \caption{$5$}
                \label{fig:}
        \end{subfigure}%
          \begin{subfigure}[t]{0.083\textwidth}
        \centering
        \captionsetup{font=small}
                \includegraphics[width=.8\textwidth]{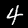}
                \caption{$4$}
                \label{fig:}
        \end{subfigure}%
          \begin{subfigure}[t]{0.083\textwidth}
        \centering
        \captionsetup{font=small}
                \includegraphics[width=.8\textwidth]{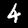}
                \caption{$4$}
                \label{fig:}
        \end{subfigure}%
            \begin{subfigure}[t]{0.083\textwidth}
        \centering
        \captionsetup{font=small}
                \includegraphics[width=.8\textwidth]{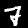}
                \caption{$7$}
                \label{fig:}
        \end{subfigure}%
          \begin{subfigure}[t]{0.083\textwidth}
        \centering
        \captionsetup{font=small}
                \includegraphics[width=.8\textwidth]{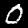}
                \caption{$0$}
                \label{fig:}
        \end{subfigure}%
          \begin{subfigure}[t]{0.083\textwidth}
        \centering
        \captionsetup{font=small}
                \includegraphics[width=.8\textwidth]{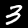}
                \caption{$3$}
                \label{fig:}
        \end{subfigure}%
        ~           
          \begin{subfigure}[t]{0.083\textwidth}
                \includegraphics[width=.8\textwidth]{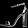}
                \caption{Boot}
                \label{fig:}
        \end{subfigure}%
          \begin{subfigure}[t]{0.083\textwidth}
        \centering
        \captionsetup{font=small}
                \includegraphics[width=.8\textwidth]{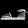}
                \caption{Sneaker}
                \label{fig:}
        \end{subfigure}%
          \begin{subfigure}[t]{0.083\textwidth}
        \centering
        \captionsetup{font=small}
                \includegraphics[width=.8\textwidth]{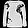}
                \caption{Pullover}
                \label{fig:}
        \end{subfigure}%
          \begin{subfigure}[t]{0.083\textwidth}
        \centering
        \captionsetup{font=small}
                \includegraphics[width=.8\textwidth]{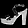}
                \caption{Boot}
                \label{fig:}
        \end{subfigure}%
          \begin{subfigure}[t]{0.083\textwidth}
        \centering
        \captionsetup{font=small}
                \includegraphics[width=.8\textwidth]{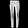}
                \caption{Trouser}
                \label{fig:}
        \end{subfigure}%
          \begin{subfigure}[t]{0.083\textwidth}
        \centering
        \captionsetup{font=small}
                \includegraphics[width=.8\textwidth]{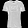}
                \caption{T-shirt}
                \label{fig:}
        \end{subfigure}%
                                                     
          \begin{subfigure}[t]{0.083\textwidth}
                   \captionsetup{font=small}
        \centering
                \includegraphics[width=.8\textwidth]{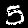}
                \caption{$I(2)$}
                \label{fig:}
        \end{subfigure}%
          \begin{subfigure}[t]{0.083\textwidth}
                   \captionsetup{font=small}
        \centering
                \includegraphics[width=.8\textwidth]{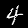}
                \caption{$I(3)$}
                \label{fig:}
        \end{subfigure}%
          \begin{subfigure}[t]{0.083\textwidth}
                   \captionsetup{font=small}
        \centering
                \includegraphics[width=.8\textwidth]{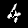}
                \caption{$I(4)$}
                \label{fig:}
        \end{subfigure}%
          \begin{subfigure}[t]{0.083\textwidth}
                   \captionsetup{font=small}
        \centering
                \includegraphics[width=.8\textwidth]{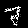}
                \caption{$D(2)$}
                \label{fig:}
        \end{subfigure}%
          \begin{subfigure}[t]{0.083\textwidth}
                   \captionsetup{font=small}
        \centering
                \includegraphics[width=.8\textwidth]{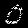}
                \caption{$D(3)$}
                \label{fig:}
        \end{subfigure}%
          \begin{subfigure}[t]{0.083\textwidth}
                   \captionsetup{font=small}
        \centering
                \includegraphics[width=.8\textwidth]{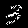}
                \caption{$D(4)$}
                \label{fig:}
        \end{subfigure}%
        ~
          \begin{subfigure}[t]{0.083\textwidth}
                   \captionsetup{font=small}
        \centering
                \includegraphics[width=.8\textwidth]{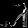}
                \caption{$I(2)$}
                \label{fig:}
        \end{subfigure}%
          \begin{subfigure}[t]{0.083\textwidth}
                   \captionsetup{font=small}
        \centering
                \includegraphics[width=.8\textwidth]{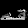}
                \caption{$I(3)$}
                \label{fig:}
        \end{subfigure}%
          \begin{subfigure}[t]{0.083\textwidth}
                   \captionsetup{font=small}
        \centering
                \includegraphics[width=.8\textwidth]{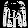}
                \caption{$I(4)$}
                \label{fig:}
        \end{subfigure}%
                  \begin{subfigure}[t]{0.083\textwidth}
                   \captionsetup{font=small}
        \centering
                \includegraphics[width=.8\textwidth]{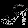}
                \caption{$D(2)$}
                \label{fig:}
        \end{subfigure}%
          \begin{subfigure}[t]{0.083\textwidth}
                   \captionsetup{font=small}
        \centering
                \includegraphics[width=.8\textwidth]{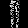}
                \caption{$D(3)$}
                \label{fig:}
        \end{subfigure}%
          \begin{subfigure}[t]{0.083\textwidth}
                   \captionsetup{font=small}
        \centering
                \includegraphics[width=.8\textwidth]{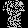}
                \caption{$D(4)$}
                \label{fig:}
        \end{subfigure}%
\caption{Randomly selected samples of original images (top row) with corresponding Ising dropout (I($N$)) or random dropout (D($N$)) (p=0.5) image (bottom row) in a network with $N$ hidden layers for MNIST and Fashion-MNIST datasets.}
\label{fig:input_ising_dropped_samples}
\vspace{-4mm}
\end{figure*}

\subsection{Technical Details of Training}
Depending on the dataset and network architecture, various hyperparamters are studied and the best values are reported. We used Adam optimizer~\cite{kingma2014adam} with adaptive learning rate starting at 0.01. No regularization method except dropout (stated if applied) was used. The maximum number of training iterations was set to 200 and early stopping was applied. The mini-batch size is set to $Q$=32.

Total number of parameters $P$ to optimize in a MLP network with $N\geq1$ hidden layers is 
\begin{equation}
P = |\mathbf{x}|\cdot|\mathbf{h}_{1}|+\sum_{i=1}^{N-1}(|\mathbf{h}_{i}|\cdot(|\mathbf{h}_{i+1}|+1))+ (|\mathbf{h}_{N}|\cdot(|\mathbf{y}|+1)) +|\mathbf{y}| 
\end{equation}
where $|h_{i}|$ is the cardinality of the layer $h_{i}$, $\mathbf{x}$ is the input vector (layer) and $\mathbf{y}$ is the output vector (layer). 

\subsection{Results and Performance Comparisons}
The performance results for three MLP network architectures, to classify MNIST images, are presented in Table~\ref{T:mnist_da}.
The results show that the proposed Ising-Dropout has competitive performance with no dropout method while accelerating the training of network by optimizing a subset of network parameters. This method can also compress the trained inference model by selecting the well-trained network weights while keeping the performance competitive. The results show that the proposed method has better dropout rate as the depth of network increases while maintaining a high performance. As an example, for an MLP with four hidden layers, the classification performance was $94.40\%$ without using dropout, where the backpropagation was performed on the entire parameters of the network and entire inference model was used for validation. This is while the proposed Ising-Dropout method achieved a classification accuracy of $90.74\%$, which is $3.66\%$ less than the no dropout method, but could drop on average $41.18\%$ of the network parameters during training. The inference model was also compressed $41.18\%$ smaller than the original network, which is approximately 36,088 parameters. This performance is much higher than random dropout of network weights.
 
The results show that applying Ising-Dropout during training and later in inference results in better classification performance, particularly for the Fashion-MNIST dataset, which is more complex than MNIST. The results for various depth of the network have similar behavior for the MNIST. However, the classification accuracy of the models is lower. There is a trade-off between performance and compression rate of the network. At $5.84\%$ lower accuracy for a 4-layer MLP, the network is $55.86\%$ smaller. 

The results also show that applying dropout on the input images can help the models achieve higher classification accuracy. Figure~\ref{fig:input_ising_dropped_samples} shows randomly selected samples from MNIST dataset and visualizes corresponding Ising-dropped image for different architectures of MLP. These examples show that the proposed method can preserve information in the input data and ignore unnecessary (e.g. background pixels) input values. The sample images show that although some pixels are removed from the digits, the shape and structure of input data is preserved.

\section{Conclusions}
\label{sec:conclusion}
Deep neural networks generally suffer from two issues, overfitting and large number of parameters to optimize. Dropout is a regularization method to improve training of deep neural networks. In this paper, we propose a dropout method based on the Ising energy, called Ising-Dropout, of the deep neural network to wisely drop input and/or hidden units from the network while training. This approach helps the network to avoid overfitting and optimize a subset of parameters in the network. 

The other application of the proposed method is to compress the trained network (inference model). The preliminary results show that there is a trade-off between network size and classification accuracy. The proposed Ising-Dropout method can reduce the number of parameters in the inference network into half while keeping the classification accuracy competitive to the original network. This approach selects nodes associated with well-trained parameters of the network for inference. This compression technique can increase inference speed while maintaining the prediction accuracy, necessary for certain applications such as mobile device and deep learning on chip. This method can also be developed for convolutional neural networks in future works.

\section{Acknowledgment}
The authors acknowledge financial support and access to the Digital Annealer (DA) of Fujitsu Laboratories Ltd. and Fujitsu Consulting (Canada) Inc.
\bibliographystyle{IEEEbib}
\bibliography{strings,mybibfile}

\end{document}